\pdfoutput=1
\documentclass[sigconf,nonacm]{acmart}

\usepackage{soul}
\usepackage{siunitx}
\usepackage{multirow}
\usepackage{subcaption}
\usepackage[dvipsnames]{xcolor}
\usepackage{booktabs}
\usepackage{colortbl}

\definecolor{bestblue}{RGB}{0,102,204}
\definecolor{acmblue}{RGB}{0, 90, 158}
\definecolor{lightgreen}{rgb}{0.5, 0.8, 0.6}
\definecolor{bleudefrance}{rgb}{0.0, 0.53, 0.74}
\definecolor{rowblue}{RGB}{232,243,255}
\definecolor{bestcell}{RGB}{255,243,205}

\newcommand{\best}[1]{\textbf{\textcolor{bestblue}{#1}}}
\newcommand{\gain}[1]{{\,\tiny\textcolor{bestblue}{$\uparrow$\!#1\%}}}
\newcommand{\second}[1]{\underline{#1}}

\AtBeginDocument{%
  }

\setcopyright{none}
\renewcommand\footnotetextcopyrightpermission[1]{}

\begin{document}

\title{MAEPose: Self-Supervised Spatiotemporal Learning for Human Pose Estimation on mmWave Video}

\author{Xijia Wei}
\email{xijia.wei.21@ucl.ac.uk}
\orcid{0000-0003-4745-6569}
\affiliation{%
  \institution{University College London}
  \city{London}
  \country{UK}
}

\author{Yuan Fang}
\email{yuan.fang.20@ucl.ac.uk}
\orcid{https://orcid.org/0009-0008-0075-5071}
\affiliation{%
  \institution{University College London}
  \city{London}
  \country{UK}
}

\author{Kevin Chetty}
\orcid{https://orcid.org/0000-0003-1616-4248}
\email{k.chetty@ucl.ac.uk}
\affiliation{%
  \institution{University College London}
  \city{London}
  \country{UK}
}

\author{Youngjun Cho}
\orcid{https://orcid.org/0000-0001-5695-0759}
\email{youngjun.cho@ucl.ac.uk}
\affiliation{%
  \institution{University College London}
  \city{London}
  \country{UK}
}

\author{Nadia Bianchi-Berthouze}
\email{nadia.berthouze@ucl.ac.uk}
\orcid{https://orcid.org/0000-0001-8921-0044}
\affiliation{%
  \institution{University College London}
  \city{London}
  \country{UK}
}

\renewcommand{\shortauthors}{Wei et al.}

\begin{abstract}
Millimetre-wave (mmWave) radar offers a more privacy-preserving alternative to RGB-based human pose estimation. However, existing methods typically rely on pre-extracted intermediate representations such as sparse point clouds or spectrogram images, where the rich spatiotemporal information naturally present in radar video streams is discarded for model learning, while such signal processing adds system complexity. In addition, existing solutions are mainly conducted in an end-to-end supervised manner without leveraging unlabelled raw video streams to learn generalized representations. In this study, we present MAEPose, a masked autoencoding-based human pose estimation approach that operates directly on mmWave spectrogram videos. MAEPose learns spatiotemporal motion-aware generalized representations from unlabelled radar video, and leverages its heatmap decoder for multi-frame pose estimation predictions. We evaluate it across three datasets based on leave-one-person-out cross-validation with rigorous statistical testing. MAEPose consistently outperforms state-of-the-art baselines by up to 22.1\% in MPJPE ($p{<}0.05$), and maintains robust accuracy under zero-shot bystander interference with only a 6.5\% error increase. Ablation studies confirm that both the pre-training and the heatmap decoder contribute substantially, while modality analysis indicates that leveraging Range-Doppler video as input achieves better pose estimation performance than Range-Azimuth or their fusion, with lower computational cost.
\end{abstract}




\maketitle

\section{Introduction}

Human pose estimation is a fundamental demand for various applications such as motion tracking, healthcare monitoring to human-computer interaction~\cite{wei2019calibrating, wei2021sensor, zhang2023survey}. While RGB camera-based methods have achieved promising accuracy and reliability, they raise significant privacy concerns~\cite{wei2023leveraging, wei2025wiprot}. In addition, RGB-based approaches fail to work under poor lighting conditions or with occlusion issues~\cite{soumya2023recent}. mmWave radar has emerged as a potential alternative, since it is considered more privacy-preserving than RGB cameras, operates through non-metal occlusions and is unaffected by lighting conditions~\cite{wei2025vomee}.

Despite growing interest, existing mmWave-based pose estimation methods typically rely on pre-extracted representations from the raw signal, such as sparse point clouds extracted via peak detection~\cite{fan2024mmdiff, zhu2024probradar}, or aggregated 2D Micro-Doppler maps with sequential or windowed temporal features~\cite{lee2023hupr, sheng2022facilitating}. However, such operations require signal pre-processing that discards the rich spatiotemporal information originally encoded in the raw mmWave spectrogram streams, while adding extra complexity to the sensing system~\cite{wei2025mmwavetryon}. Moreover, existing systems are mainly trained in an end-to-end supervised manner for task-specific purposes, rather than learning a generalized representation from the radar signal.

Our key insight is that mmWave sensing is inherently spatiotemporal: continuous chirp transmissions produce sequences of spectral images, such as Range-Doppler (RD) and Range-Angle (RA), that are naturally represented as a video formulation. This motivates the following research question: can we design a model that directly processes mmWave video streams without extra signal processing (e.g., task-specific frequency filtering, noise removal or temporal feature aggregation), and learns a generalized spatiotemporal representation in a self-supervised manner over the unlabelled mmWave videos and adapts to the pose estimation task? To address this, we propose \textsc{\textbf{MAEPose}}, a \textbf{M}asked \textbf{A}uto\textbf{E}ncoding-based \textbf{Pose} estimation framework that operates directly on mmWave spectrogram videos. MAEPose learns spatiotemporal motion-aware generalized representations from unlabelled radar video via masked autoencoding, followed by a multi-frame heatmap decoder for pose estimation.

To the best of our knowledge, no prior work has directly processed mmWave spectrogram as video streams with self-supervised spatiotemporal learning for pose estimation. We summarize the following contributions:

\begin{figure*}[t]
    \centering
    \includegraphics[width=\textwidth]{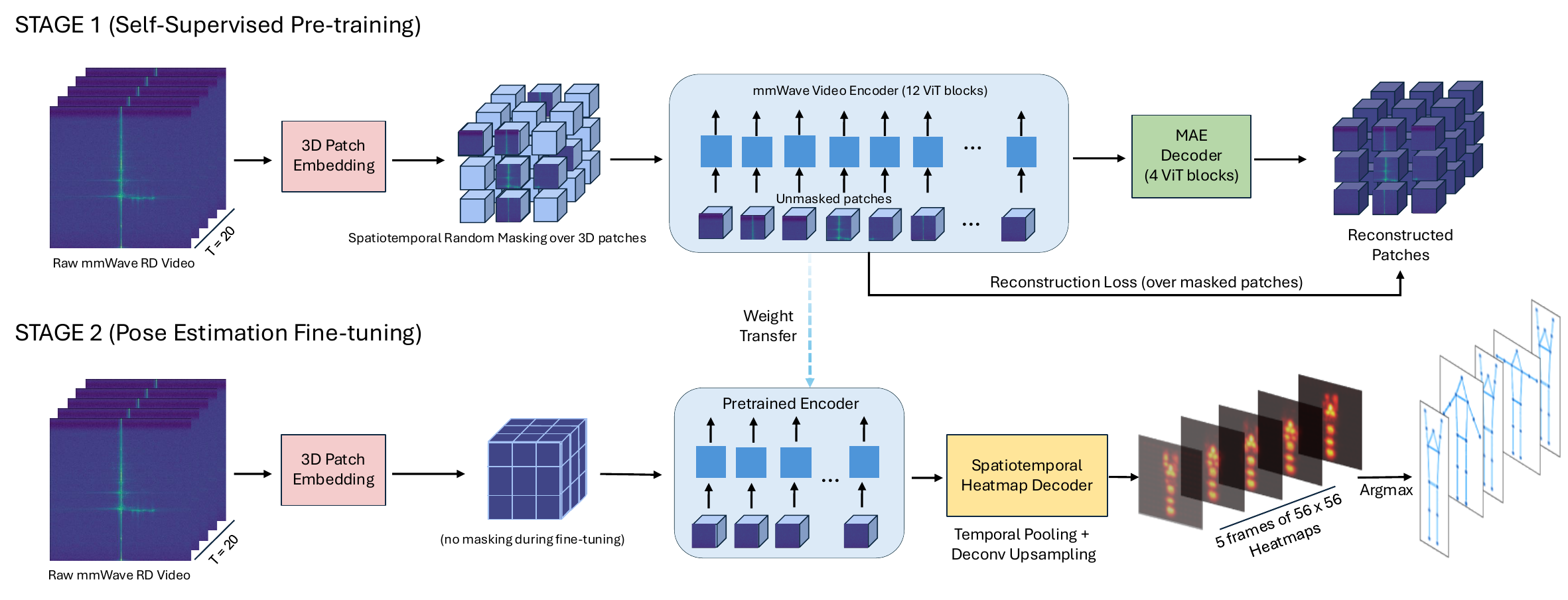}
    \caption{Overview of the MAEPose architecture. The model processes sequences of Range-Doppler radar spectrograms through its video Transformer encoder, followed by a spatial heatmap-based pose decoder for multi-frame human pose estimation.}
    \label{fig:MAEPose_model}
\end{figure*}

\begin{enumerate}
    \item We propose MAEPose, a novel self-supervised mmWave video-based pose estimation framework. MAEPose operates directly on raw radar spectrogram video streams through a video Vision Transformer encoder that learns spatiotemporal representations via masked autoencoding pre-training, followed by a multi-frame heatmap decoder that preserves spatial correspondence between encoder features and joint locations, rather than collapsing the learnt features directly for skeleton joint regression.

    \item We conduct extensive experiments across three mmWave pose estimation datasets with leave-one-person-out cross-validation and statistical analysis. MAEPose consistently outperforms state-of-the-art supervised baselines across all datasets, and shows robust performance under zero-shot bystander interference with only a 6.5\% error increase compared to its controlled-setting performance.

    \item We systematically investigate the impact of varying input modalities (RD, RA, and dual-stream fusion) on pose estimation performance. Results indicate that RD alone offers rich motion information for pose estimation while using RA or dual-stream fusion brings no significant improvements or even performance degradation.
    
    \item Through ablation studies, we show that both the MAE pre-training and the multi-frame heatmap decoder contribute substantially to MAEPose's performance gains, reducing pose error by up to 41\% over training from scratch and up to 45\% over conventional MLP/GCN regression heads.
\end{enumerate}

\section{Related Work}
\label{sec:related_work}

\subsection{mmWave-Based Human Sensing}
\label{sec:rw_pose}
mmWave radar has gained increasing attention for human sensing thanks to its privacy-preserving and non-intrusive characteristics, serving as an alternative sensing modality to RGB cameras or wearable sensors that raise privacy concerns, fail under occlusion, or restrict users' movement. mmWave signals can be represented in various formulations, including raw radar cubes~\cite{zhao2023cubelearn, fang2025cubedn}, spectrograms~\cite{lee2023hupr, rahman2023radar}, and point clouds~\cite{fan2024mmdiff, zhu2024probradar}. Depending on the tasks, different mmWave representations are chosen for different sensing purposes such as activity recognition, and mesh reconstruction. We introduce the existing works grouped by their input representation.

\noindent\textbf{Point-cloud-based methods.} 
Works including mmDiff~\cite{fan2024mmdiff}, ProbRadarM3F~\cite{zhu2024probradar} and milliMamba~\cite{kini2026millimamba} take the radar point clouds to reconstruct the static skeletons on a single-frame level. These works adopt generative approaches such as diffusion models~\cite{ho2020denoising} to reconstruct human skeleton or human mesh. However, the number of point clouds is limited by the resolution of the radar hardware specs~\cite{zhu2024probradar}. This makes the point clouds inherently sparse and noisy, limiting the ability for describing fine-grained motion details required for precise skeleton reconstruction. In addition, extracting radar point clouds relies on additional CFAR~\cite{cfar} operation over the raw RAD cube, adding preprocessing complexity to the system pipeline.

\noindent\textbf{Spectrogram-based methods.} In contrast, mmWave spectrograms preserve richer information. HuPR~\cite{lee2023hupr} uses 3D CNN (convolutional neural network) and GCN (graph convolutional network) for skeleton reconstruction; CubeLearn~\cite{zhao2023cubelearn} introduces a complex-valued 3D CNN backbone to process the radar cube into image-like feature representations, then leverages LSTM layers for posture recognition. However, these two methods are trained end-to-end in a fully supervised manner, without the potential to leverage the unlabelled data streams naturally produced by continuous radar sensing for learning the generalized spatiotemporal representations. %Therefore, MAEPose offers a straightforward yet elegant approach that operates directly on mmWave video streams without extra data processing with unlabelled spatiotemporal representation learning capability.

\subsection{Masked Autoencoding-Based Self-Supervised Learning}
\label{sec:rw_mae}
Masked autoencoding approach has become a dominant SSL paradigm. The concept originated from masked language modelling in BERT~\cite{devlin2019bert}, where part of the textual tokens is randomly masked and the pretext task aims to reconstruct the masked tokens. He et al.~\cite{he2022mae} brought this idea to visual modality by applying masked reconstruction over image patches~\cite{dosovitskiy2020image}. It was then extended to video representation by masking over spatiotemporal patches across frames~\cite{feichtenhofer2022masked, tong2022videomae, wang2023videomaev2}. In terms of the pose estimation task, PoseFormer~\cite{zheng20213d} leverages ViT (vision transformer) to extract RGB image features for pose estimation, while ViTPose~\cite{xu2022vitpose} is built on a ViT backbone and adapts MAE pre-training, which achieves the SOTA pose estimation performance on RGB-image-based input. Beyond textual and visual modalities, AudioMAE~\cite{huang2022audiomae} applied MAE on audio spectrograms (as image formulation), demonstrating that MAE can also understand acoustic information conveyed through FFT spectrogram representations. 

Despite its widespread adoption for SSL-based training
across modalities (textual: natural language understanding;
visual: RGB-based pose estimation; acoustic: audio content
recognition), to the best of our knowledge, no prior work has
built upon MAE to directly learn generalized spatiotemporal
representations from unlabelled mmWave spectrogram video
for pose estimation. This motivates the design of our proposed MAEPose.

\section{MAEPose Framework}

MAEPose is a masked-autoencoding-based human pose estimation framework that directly processes mmWave spectrogram video as input through its pre-trained feature encoder and predicts multi-frame skeleton joint heatmaps via its spatiotemporal heatmap decoder. 2D skeleton joint values are calculated via argmax over the heatmaps. Training MAEPose contains two stages, as shown in Figure~\ref{fig:MAEPose_model}.

In \textbf{Stage 1 (Self-Supervised Pretraining)}, a video-based masked reconstruction task is used for training MAEPose to learn the spatiotemporal representation without the need for human pose annotations. The task aims to train MAEPose to reconstruct the mmWave video spectrogram patches given a partially masked mmWave video. During pre-training, a video ViT (Vision Transformer) encoder extracts the spatiotemporal representation from the unmasked mmWave video patches, while the decoder ViT receives such embedded features and reconstructs the masked patches. The reconstruction loss is calculated as the mean squared error (MSE) over the masked patches, measuring the similarity between the ground truth masked mmWave video patches and MAEPose's reconstructed counterparts.

In \textbf{Stage 2 (Supervised Finetuning)}, MAEPose is fine-tuned based on annotated mmWave video data to adapt to the downstream task of human pose estimation. We keep MAEPose's pre-trained encoder as the spatiotemporal feature extractor, while the reconstruction decoder is discarded. A lightweight \emph{multi-frame heatmap-based pose decoder} (\S\ref{sec:heatmap_decoder}) is added, consisting of a Conv3d layer followed by spatial upsampling, to decode the extracted features into per-joint 2D heatmaps across multiple frames. An argmax calculation over MAEPose's generated heatmaps converts them to 2D skeleton joint values.

\subsection{Stage 1: Spatiotemporal MAE Pre-training}
\subsubsection{mmWave Video Patch Embedding}
MAEPose first processes the raw mmWave video inputs into video patches. The patch embedder operates 3D convolutional embedding with its kernel size and stride step of $2 \times 16 \times 16$. MAEPose processes a 20-frame mmWave video clip as input, which is patched into $10 \times 14 \times 14 = 1{,}960$ spatiotemporal tokens, with each embedding dimension of 384. This design is inspired by the standard video masked autoencoder of~\cite{feichtenhofer2022masked}.

\subsubsection{Video Patch Random Masking}
After the mmWave video clip is patched into spatiotemporal tokens, we randomly mask 90\% of the tokens in spacetime, so that only the remaining 10\% (${\sim}196$ tokens) will be received by the ViT component. This high masking ratio is motivated by the fact that mmWave video streams are highly sparse and noisy with high information redundancy in spacetime, where consecutive frames share similar backgrounds while only a small portion of the pixels per frame contain human bodily movement features. A high masking ratio makes the reconstruction task significantly challenging, therefore forcing MAEPose to learn the human-motion-related representation from the sparse unmasked mmWave video information.

\subsubsection{MAEPose Video Encoder}
MAEPose adopts a standard ViT (Vision Transformer) as its video encoder backbone. It consists of 12 ViT blocks, with 384 dimensions and 6 attention heads. The visible mmWave video tokens (the remaining 10\%) are added with position embeddings to preserve their original spatiotemporal information, and then fed into the encoder for feature extraction.

\subsubsection{MAEPose Video Reconstruction Decoder}
MAEPose, during the pre-training stage, consists of a lightweight ViT decoder (4 blocks, 512 dimensions, 16 attention heads). The decoder receives the encoded unmasked token features, concatenated with learnable mask tokens at those positions where original patches are being masked. Therefore, the decoder understands the spatiotemporal relationship when reconstructing those masked patches given their encoded features together with the masked tokens. The reconstruction loss is calculated based on the mean squared error between ground-truth patches and their reconstructed counterparts over masked positions only. The hypothesis is that through the reconstruction pre-training, if the decoder can reconstruct the masked patches based on its encoded features, it indicates that the encoder has learnt good spatiotemporal representation capability to encode the highly sparse and redundant mmWave spectrogram video streams in spacetime.

\subsection{Stage 2: Pose Estimation Fine-tuning}
In the fine-tuning stage, MAEPose's reconstruction decoder is discarded, while its encoder is transferred to the downstream pose estimation task and used as the spatiotemporal feature extractor. In addition, a heatmap-based pose decoder is added for processing the extracted features to generate multi-frame heatmaps (\S\ref{sec:heatmap_decoder}). Eventually, skeleton keypoints are derived via a simple argmax calculation based on the generated heatmaps; this step is not involved in the model fine-tuning.

\subsubsection{Pre-trained Encoder Transfer}
\label{sec:encoder_transfer}
MAEPose initializes the 3D patch embedding and the ViT-based encoder's parameters from the pre-trained weights, learnt during the reconstruction task. During the fine-tuning, the encoder processes the entire mmWave video clip input instead of applying random masking, which is only used in pre-training. Since the encoder has gained the spatiotemporal understanding capability from its pre-training, it can automatically extract the generalized key information in spacetime while discarding the sparse and noisy visual factors appearing in the video. To preserve the pre-trained knowledge while allowing the projection head, the pose decoder, to understand the alignment between the generalized extracted knowledge and the downstream heatmap generation task, a layer-wise learning rate decay is applied, a common technique for masked auto-encoding-based model fine-tuning.

\subsubsection{Multi-frame Heatmap-Based Pose Decoder}
\label{sec:heatmap_decoder}
One of the novel designs in MAEPose is the multi-frame heatmap pose decoder. Since MAEPose is proposed for mmWave video understanding, it naturally processes information in spacetime. When applied to pose estimation, this enables the model to leverage the spatiotemporal representation for predicting poses across multiple frames simultaneously rather than treating each frame independently, which loses the complementary information conveyed across frames.

MAEPose adopts a modern pose estimation framework design concept that predicts heatmaps, and extends it to multi-frame heatmap outputs, rather than directly predicting coordinate regression (e.g., $(x, y)$ values for each joint). The lightweight decoder consists of a 3D convolutional layer with kernel $(3, 1, 1)$ and stride $(2, 1, 1)$. It decodes the spatiotemporal features into dense 2-dimensional probability maps where pixels represent the likelihood of a human body joint being present at a corresponding spatial position on the heatmap. Specifically, it decodes the extracted feature by upsampling the spatial resolution from $14 \times 14$ to $56 \times 56$ ($14 \rightarrow 28 \rightarrow 56$). A $1 \times 1$ convolution produces $K{=}13$ heatmap channels with the shape of $(B, 5, 13, 56, 56)$, indicating 13 body joint corresponding heatmaps.

MAEPose operates within the same data formulation that both its inputs (mmWave visual streams) and outputs (heatmaps) are in visual representation. This design avoids the spatially encoded features being collapsed into a series of single human-body-joint vector, which discards the learnt rich representation during pose estimation fine-tuning. Instead, learning to generate the visual-represented heatmaps, as a more complex task, forces the model to preserve learnt rich information for understanding pose-related representation.

Regarding the fine-tuning ground truth, heatmaps are generated by placing 2D Gaussians at each joint's coordinates, originally extracted from RGB frames during data collection. The loss is a weighted MSE between predicted and target heatmaps.

\subsubsection{Heatmap to Skeleton}
\label{sec:heatmap2skeleton}
To derive the skeleton keypoints values, joint coordinates are extracted from the predicted heatmap $\mathbf{H}_k \in \mathbb{R}^{H' \times W'}$. For each joint $k$, it is calculated via the argmax:
\begin{equation}
    (\hat{u}_k, \hat{v}_k) = \arg\max_{(i,j)} \mathbf{H}_k(i, j)
\end{equation}
\begin{equation}
    \hat{x}_k = \frac{\hat{v}_k}{W'}, \quad \hat{y}_k = \frac{\hat{u}_k}{H'}
\end{equation}
where $(\hat{x}_k, \hat{y}_k) \in [0, 1]^2$ are the normalized joint coordinates in the image plane. This calculation returns multi-frame skeleton outputs of 5 target frames $\times$ 13 joints $\times$ 2 coordinates per video input sample.

\section{Experiments}

\subsection{Data Collection}

\begin{figure}[tb]
  \centering
  \begin{subfigure}[t]{0.463\linewidth}
    \centering
    \includegraphics[width=\linewidth]{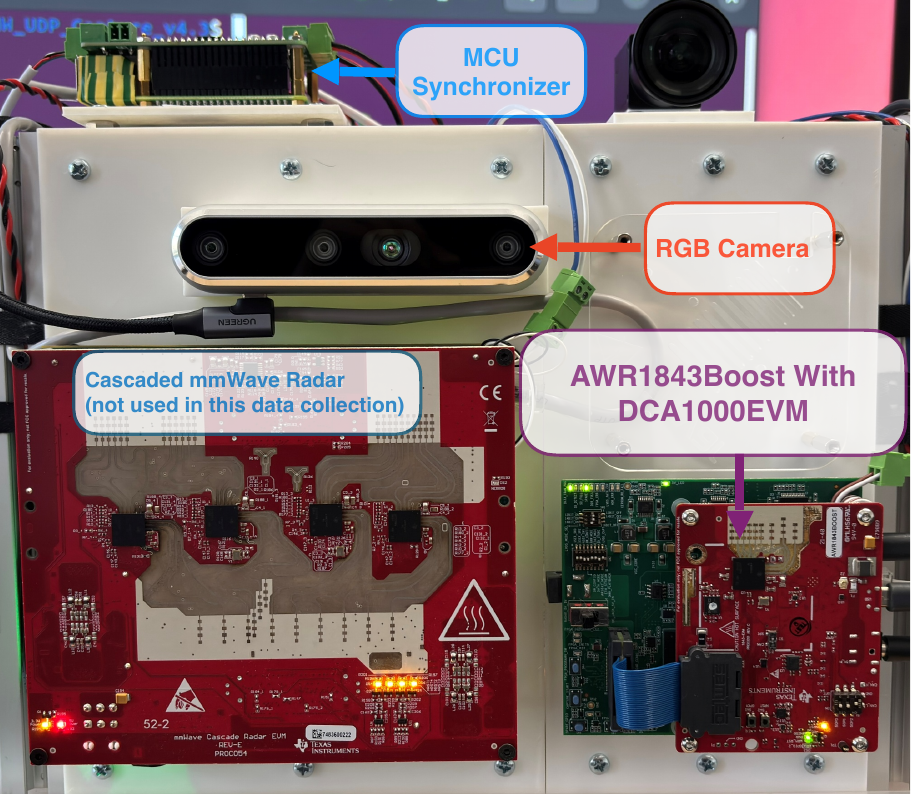}
    \caption{Data Collection Hardware}
    \label{fig:sensors}
  \end{subfigure}
  \hfill
  \begin{subfigure}[t]{0.48\linewidth}
    \centering
    \includegraphics[width=\linewidth]{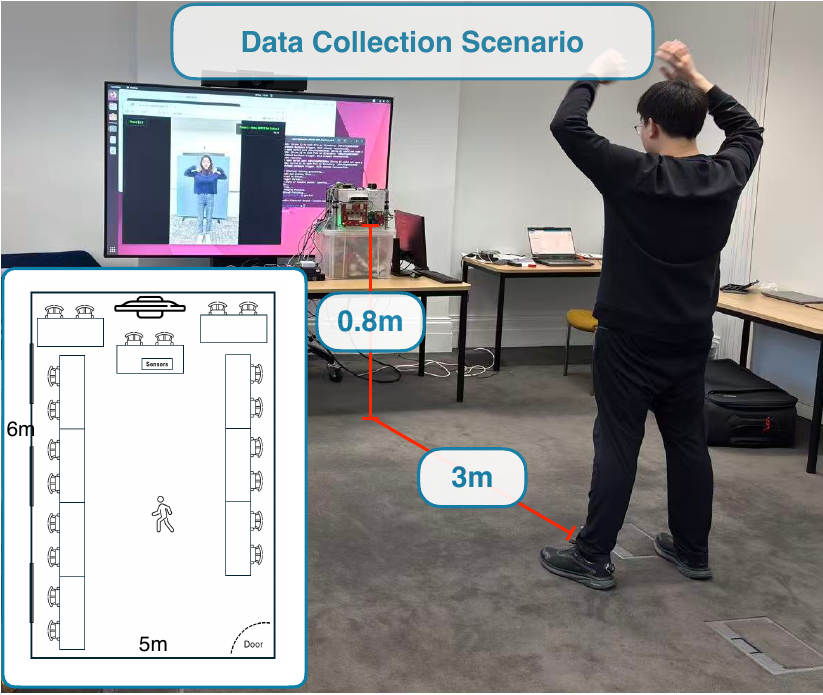}
    \caption{Capture Scenario}
    \label{fig:room}
  \end{subfigure}
  \caption{The multi-sensory data collection platform and the capture scenario with room layout.}
  \label{fig:setup}
\end{figure}

Figure~\ref{fig:setup} shows our multimodal data collection settings~\cite{wei2025vomee}, including the data collection hardware (Figure~\ref{fig:sensors}) and a diagram of the capture scene (Figure~\ref{fig:room}). The hardware platform consists of a Texas Instruments AWR1843BOOST single-chip radar with a DCA1000EVM capture card for capturing mmWave signals, and an Intel Realsense D455 camera for RGB recording. The mmWave radar operates at 77 GHz start frequency, 65.998~MHz/$\mu$s slope, 4800~ksps sampling rate, 256 ADC samples, and 255 chirp loops. It corresponds to a 0.0426~m range resolution and 0.0477~m/s Doppler resolution. All sensors are synchronized using a hardware trigger generated by a microcontroller unit (MCU), ensuring a uniform sampling rate of 10 Hz.

\subsection{Datasets}
We evaluate MAEPose across three datasets collected from multiple indoor environments with distinct participants and action vocabularies. Table~\ref{tab:datasets} summarizes the dataset characteristics. The ground truth consists of 13 2D body skeleton keypoints, extracted from the time-aligned RGB frames using ViTPose~\cite{xu2022vitpose}, where points 0--12 represent the nose, left shoulder, left elbow, left wrist, right shoulder, right elbow, right wrist, left hip, right hip, left knee, right knee, left ankle and right ankle, respectively.

\begin{table}[tb]
\centering
\caption{Overview of the Datasets}
\label{tab:datasets}
\renewcommand{\arraystretch}{0.85}
\setlength{\tabcolsep}{3pt}
\resizebox{\columnwidth}{!}{%
\begin{tabular}{lccccccl}
\toprule
\textbf{Dataset} & \textbf{Pers.} & \textbf{Acts.} & \textbf{Skel.} & \textbf{Frames} & \textbf{Time (m)} & \textbf{Modality} & \textbf{Room} \\
\midrule
\textit{mmTryOn} & 8 & 11 & 13 & 40.6K & 67.6 & Raw, RD, RA & Lab (12m\textsuperscript{2})\\
\textit{mmMove}  & 9 & 10 & 13 & 43.9K & 73.2 & Raw, RD, RA & Studio (30m\textsuperscript{2})\\
\textit{mmYoga}  & 9 &  9 & 13 & 23.7K & 39.5 & Raw, RD, RA & Studio (30m\textsuperscript{2})\\
\bottomrule
\end{tabular}%
}
\vspace{2pt}
\parbox{\columnwidth}{\tiny
\textit{mmTryOn}: P1--P8, A1--A11 (try-on gestures), collected in a lab room.
\textit{mmMove}: P9--P17, A1--A10 (full-body movements).
\textit{mmYoga}: P9--P17, A12--A20 (yoga poses); collected at the same studio room \& participant groups as \textit{mmMove}. 13 body joints selected from 17 COCO keypoints from ViTPose: nose, shoulders, elbows, wrists, hips, knees, and ankles --- excluding eyes and ears. Time counts only active action durations. All datasets contain raw mmWave data, supporting the extraction of Range-Doppler \& Range-Azimuth spectrograms.}
\end{table}

\textbf{\textit{mmTryOn.}} mmTryOn is built upon~\cite{wei2025mmwavetryon} by extracting skeleton ground truth from RGB frames. mmTryOn is a full-body activity recognition dataset including clothes try-on gestures, captured by the mmWave-based smart mirror. Eight participants (females aged from 20-50) performed 11 try-on gestures (A1--A11) in a laboratory room ($\sim$12\,m$^2$), including shoulder roll, sleeve adjust, stepping, arm raise, torso twist, chest expand, arm flap, head turn, hand slide, arm swing, and forearm lift.

\textbf{\textit{mmMove.}} Collected in a larger studio room (over $\sim$30\,m$^2$). Nine participants (male and female aged from 20 - 40) performed 10 actions (A1--A10). The action vocabulary largely overlaps with \textit{mmTryOn}. In addition, mmMove contains a real-world sensing phase where the target person is performing actions while other persons are walking randomly surrounding the room. This enables us to evaluate our model's robustness under zero-shot real-world situations where irrelevant moving objects appear in the sensing environment. %It enables cross-dataset comparisons despite different rooms and participants.

\textbf{\textit{mmYoga.}} Inspired by yoga full-body gestures, which are challenging and hard to perform. This dataset aims to increase performing difficulties, reducing the repeatable body patterns appearing in the dataset. During collection, we observed that participants found it hard to repeat gestures in a similar pattern within the same type of actions, while body torso was less controlled due to the hardness of the gestures. We used the same studio room and recruited the same group of participants as in \textit{mmMove}. \textit{mmYoga} includes 9 entirely different actions (A12--A20): crescent side stretch, standing heart opener, gate pose reach, dynamic knee march, shoulder opening flow, revolved crescent lunge, skater's balance flow, warrior chest opener, and reverse warrior reach. These actions involve more complex whole-body coordination, testing MAEPose's ability to handle motion pattern variations present in the dataset.

The three datasets span two rooms, 17 unique individuals including male and female aged from 20 to 50, height from 155 to 190 cm, and 21 distinct actions covering full-body movements, try-on gestures, and yoga exercise gestures under distinctive scenarios, totalling over 108,200 frames ($\sim$180 minutes) of synchronized radar and skeleton samples.

\subsection{Data Pre-Processing}
\label{sec:sensing_principle}

\begin{figure}[h]
  \centering
  \includegraphics[width=0.65\linewidth]{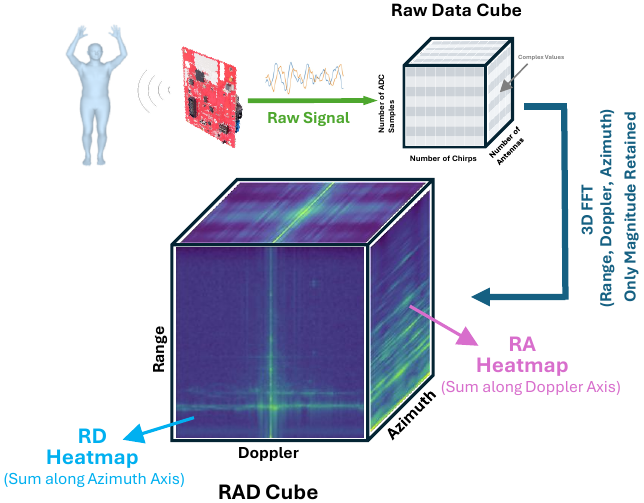}
  \caption{Radar signal processing pipeline from raw signals to the Range–Azimuth–Doppler (RAD) cube via FFT.}
  \Description{Diagram showing the radar signal processing pipeline from raw signals to the Range–Azimuth–Doppler (RAD) cube.}
  \label{fig:signal_processing}
  \vspace{-1em}
\end{figure}

mmWave raw radar signals encode physical attributes of the targets. Specifically, the propagation delay of the reflected signal provides range information, the phase variation across multiple chirps reveals Doppler velocity, and the phase difference across multiple receive antennas indicates angle-of-arrival estimation~\cite{mmwave_principle}. Through these measurements, mmWave data represents spatial and motion information of targets, including range, radial velocity, and azimuth.

The returned raw radar signals are represented in complex I/Q form and require further signal processing, such as Fast Fourier Transform (FFT), to recover target information (i.e., range, Doppler, and azimuth)~\cite{fang2025cubedn}. As illustrated in Figure~\ref{fig:signal_processing}, these measurements are transformed into a three-dimensional Range–Azimuth–Doppler (RAD) cube via FFT, where each dimension corresponds to range, Doppler velocity, and azimuth, respectively. Each bin in the RAD cube represents the reflected signal intensity from a specific spatial location and motion state, indicating the presence and strength of potential targets. By aggregating the RAD cube along one dimension to obtain two-dimensional representations, we can derive the Range–Doppler (RD) and Range–Azimuth (RA) heatmaps. The RD representation captures the motion patterns of targets, while the RA representation provides spatial distribution and angle-of-arrival information.

\subsection{Evaluation Metrics}
\label{subsec:eval_metrics}
We use Leave-One-Person-Out (LOPO) cross-validation for all experiments across three datasets. In each fold, one person's data is consistently left out during both pre-training and fine-tuning stages, while the remaining persons are split into training and validation sets (splitting seed=42 for reproducibility).

We report two metrics which are commonly used for pose estimation:
\begin{itemize}
    \item \textbf{MPJPE} (Mean Per-Joint Position Error, in metres, $\downarrow$): the average Euclidean distance between the predicted joints (argmax from generated heatmap) and ground-truth joint positions of all joints and frames per sample.
    \item \textbf{PCK@0.05} (Percentage of Correct Keypoints at 5\,cm, $\uparrow$): the proportion of predicted joints (argmax from generated heatmap) that have precision within a 5\,cm Euclidean distance threshold of the ground truth joints.
\end{itemize}

For statistical testing, we first apply the Friedman omnibus test (non-parametric repeated-measures ANOVA)~\cite{friedman1937use} across all methods. When Friedman's $p < 0.05$, we perform post-hoc pairwise comparisons by the following steps: we check the normality of the per-fold values for each method using the Shapiro-Wilk test~\cite{razali2011power} ($\alpha{=}0.05$) as a practical proxy for the normality assumption of paired differences; when all methods pass the check, we then apply paired $t$-tests and report Cohen's $d$ as the effect size. Otherwise, we use Wilcoxon signed-rank tests (two-tailed)~\cite{woolson2007wilcoxon} and report the effect size $r{=}|Z|/\sqrt{2n}$. Bonferroni correction~\cite{weisstein2004bonferroni} is applied to pairwise $p$-values ($p_{\text{adj}}{=}\min(1, k{\cdot}p_\text{raw})$, $k$ = number of pairwise comparisons). Effect size categories follow Cohen's convention~\cite{cohen1988statistical}: $r$ of $0.1/0.3/0.5$ or $d$ of $0.2/0.5/0.8$ indicates small/medium/large effects.

\subsection{SOTA Baselines}
\label{sec:baselines}
Existing methods either process radar frames independently, or aggregate temporal information into an image-like representation that can be processed by an image-based feature extractor for pose estimation. No prior work directly operates on the mmWave spectrogram videos. For a fair comparison, we adapt the selected SOTA (State-of-the-Art) baselines by replacing their original image-based feature extractors with the same Conv3d backbone used in our MAEPose model, enabling all models to process mmWave video streams under a unified input formulation. Apart from the input adaptation, the rest of each baseline architecture retains its original design.

\textbf{CubeLearn.} CubeLearn~\cite{zhao2023cubelearn} is a representative CNN+LSTM architecture commonly adopted for mmWave-based sensing systems~\cite{rahman2023radar}. The baseline consists of a Conv3d layer to process the video streams, followed by a bidirectional LSTM layer that captures temporal dependencies across frames, and a multi-layer MLP head that regresses joint coordinates.

\textbf{PoseFormer.} PoseFormer is a transformer-based framework, originally proposed for RGB image-based pose estimation~\cite{zheng20213d}. We adapt it to the radar domain by replacing its Conv2d image-based feature extractor with a Conv3d backbone for processing radar video inputs. The extracted features are fed into a Transformer-backbone-based encoder, the same backbone used in ours, followed by an MLP head that regresses the joint coordinates.

\textbf{HuPR.} HuPR is an mmWave spectrogram image based pose estimation framework. We adapt it with the same Conv3d input layer to process mmWave video, and pass the extracted features to an attention mechanism followed by an MLP head for skeleton prediction.

\subsection{Implementation Details}
All models, at the input level, are standardized via the Conv3d layer adaptation, to process mmWave video input of a sequence of 20 mmWave spectrogram frames per sample.

All experiments are conducted with leave-one-person-out (LOPO) cross-validation. Per-fold data splitting strategy remains identical across all models' training and evaluation, with random seed fixed to 42 for
reproducibility. Baselines are trained for 100 epochs with early stopping of 10 epochs. For MAEPose, we pre-train for 100 epochs with 90\% spatiotemporal masking ratio. Fine-tuning runs for 100 epochs with weight decay 0.05, and early stopping patience of 10 epochs, which follows the pre-training/fine-tuning strategy from~\cite{bao2021beit, feichtenhofer2022masked, tong2022videomae}.

All trainings are conducted on a workstation with an AMD Threadripper 5955WX CPU, 2$\times$ NVIDIA RTX~4090 GPUs, and 128~GB RAM, running on Ubuntu~22.04 with PyTorch~2.5 and CUDA~12.4. Regarding the inference latency, all models achieve real-time inference on a single RTX 4090 with over 110 FPS of inference speed, well above the 10 Hz mmWave sensing frequency.

\section{Results}
Results are reported using mean$\pm$standard deviation (LOPO cross-validation) with statistical results of all models across three datasets. We first report the performances of MAEPose and baselines under each dataset, followed by action-level pose estimation performance comparison. We then report model performances under different types of mmWave video inputs including Range-Doppler, Range-Azimuth, and a fusion of RD and RA as dual inputs, to investigate the impacts of modality variations. Regarding the ablation study, we compare MAEPose performances of i) contribution from pre-training, through comparing the model with or without pre-training weights initialized, and ii) contribution from downstream projection head architecture, through comparing the heatmap decoder with argmax skeleton calculation, GCN (Graph Convolutional Networks) and MLP (Multilayer Perceptron) regression heads.

\subsection{Main Comparison}
Table~\ref{tab:main_results} presents a comparison between our proposed MAEPose and all baselines across three datasets. We focus on Range-Doppler (RD) results in this section while cross-modality comparisons are reported in \S\ref{sec:modality_study}.

\begin{table*}[t]
\centering
\caption{Pose estimation results across three datasets (LOPO cross-validation, mean$\pm$std). MPJPE $\downarrow$: lower is better; PCK $\uparrow$: higher is better. \best{Blue bold}: best result with improvement over \second{second best} (underlined). $p$-value: Bonferroni-adjusted MPJPE-based pairwise test vs MAEPose.}
\label{tab:main_results}
\footnotesize
\renewcommand{\arraystretch}{1.1}
\begin{tabular*}{\textwidth}{@{\extracolsep{\fill}}l ccc ccc ccc @{}}
\toprule
& \multicolumn{3}{c}{\textit{mmTryOn} (8-fold)} & \multicolumn{3}{c}{\textit{mmMove} (9-fold)} & \multicolumn{3}{c}{\textit{mmYoga} (9-fold)} \\
\cmidrule(lr){2-4} \cmidrule(lr){5-7} \cmidrule(lr){8-10}
Method & MPJPE (m)\,$\downarrow$ & PCK (\%)\,$\uparrow$ & $p$-value & MPJPE (m)\,$\downarrow$ & PCK (\%)\,$\uparrow$ & $p$-value & MPJPE (m)\,$\downarrow$ & PCK (\%)\,$\uparrow$ & $p$-value \\
\midrule
\addlinespace[2pt]
\multicolumn{10}{@{}l@{}}{\textbf{\textit{RD}}} \\
\addlinespace[2pt]
CubeLearn       & \second{.0344{\tiny$\pm$.005}} & 82.0{\tiny$\pm$7}           & .0352 & .0387{\tiny$\pm$.009}          & 75.4{\tiny$\pm$12}          & .0352 & .0473{\tiny$\pm$.008}          & 67.0{\tiny$\pm$10}          & .0231 \\
PoseFormer     & \second{.0344{\tiny$\pm$.006}} & \second{83.3{\tiny$\pm$7}}  & .0345 & \second{.0385{\tiny$\pm$.010}} & \second{76.4{\tiny$\pm$16}} & .0231 & \second{.0447{\tiny$\pm$.005}} & \second{70.3{\tiny$\pm$7}}  & .0348 \\
HuPR           & .0365{\tiny$\pm$.005}          & 79.9{\tiny$\pm$7}           & .0352 & .0388{\tiny$\pm$.010}          & 75.4{\tiny$\pm$16}          & .0231 & .0487{\tiny$\pm$.007}          & 64.9{\tiny$\pm$8}           & .0231 \\
MAEPose (Ours) & \best{.0268{\tiny$\pm$.004}} \gain{22.1} & \best{91.1{\tiny$\pm$5}} \gain{9.4}  & --- & \best{.0307{\tiny$\pm$.008}} \gain{20.3} & \best{85.3{\tiny$\pm$11}} \gain{11.7} & --- & \best{.0419{\tiny$\pm$.005}} \gain{6.3}  & \best{75.4{\tiny$\pm$5}} \gain{7.2}  & --- \\
\midrule
\addlinespace[2pt]
\multicolumn{10}{@{}l@{}}{\textbf{\textit{RA}}} \\
\addlinespace[2pt]
CubeLearn       & \second{.0448{\tiny$\pm$.007}} & \second{71.3{\tiny$\pm$11}} & .0352 & \second{.0435{\tiny$\pm$.010}} & 69.0{\tiny$\pm$20}          & .0752 & .0546{\tiny$\pm$.011}          & 56.0{\tiny$\pm$19}          & .0348 \\
PoseFormer     & .0453{\tiny$\pm$.007}          & 71.5{\tiny$\pm$10}          & .0352 & .0443{\tiny$\pm$.018}          & \second{74.9{\tiny$\pm$12}} & .7072 & \second{.0551{\tiny$\pm$.011}} & \second{59.6{\tiny$\pm$12}} & .0231 \\
HuPR           & .0470{\tiny$\pm$.007}          & 67.5{\tiny$\pm$12}          & .0352 & .0481{\tiny$\pm$.012}          & 64.5{\tiny$\pm$19}          & .0626 & .0554{\tiny$\pm$.008}          & 54.4{\tiny$\pm$14}          & .0229 \\
MAEPose (Ours) & \best{.0345{\tiny$\pm$.005}} \gain{23.0} & \best{85.6{\tiny$\pm$5}} \gain{19.2} & --- & \best{.0381{\tiny$\pm$.007}} \gain{12.4} & \best{79.1{\tiny$\pm$9}} \gain{7.9}   & --- & \best{.0527{\tiny$\pm$.010}} \gain{4.4}  & \best{65.2{\tiny$\pm$10}} \gain{9.4} & --- \\
\midrule
\addlinespace[2pt]
\multicolumn{10}{@{}l@{}}{\textbf{\textit{Dual}}} \\
\addlinespace[2pt]
CubeLearn       & .0372{\tiny$\pm$.006}          & 80.3{\tiny$\pm$7}           & .0352 & \second{.0346{\tiny$\pm$.008}} & 81.6{\tiny$\pm$7}           & .0852 & .0497{\tiny$\pm$.006}          & 63.2{\tiny$\pm$9}           & .0124 \\
PoseFormer     & \second{.0358{\tiny$\pm$.006}} & 79.7{\tiny$\pm$9}           & .0348 & .0359{\tiny$\pm$.007}          & \second{80.5{\tiny$\pm$14}} & .0231 & \second{.0451{\tiny$\pm$.004}} & \second{69.9{\tiny$\pm$5}}  & .1106 \\
HuPR           & .0362{\tiny$\pm$.005}          & \second{80.3{\tiny$\pm$7}}  & .0352 & .0350{\tiny$\pm$.006}          & 80.4{\tiny$\pm$5}           & .0984 & .0460{\tiny$\pm$.004}          & 69.5{\tiny$\pm$5}           & .0333 \\
MAEPose (Ours) & \best{.0265{\tiny$\pm$.004}} \gain{26.0} & \best{91.2{\tiny$\pm$5}} \gain{10.8} & --- & \best{.0313{\tiny$\pm$.007}} \gain{9.5}  & \best{86.1{\tiny$\pm$9}} \gain{13.6}  & --- & \best{.0433{\tiny$\pm$.005}} \gain{4.3}  & \best{74.3{\tiny$\pm$4}} \gain{6.3}  & --- \\
\bottomrule
\end{tabular*}
\end{table*}

\textbf{\textit{mmTryOn:}} MAEPose achieves its strongest performance for the RD-based mmWave video stream inputs compared to all baselines. It reaches a 0.0268\,m MPJPE with 91.1\% PCK@5cm, demonstrating a 22.1\% MPJPE error reduction compared with the best-performing baseline (CubeLearn: 0.0344\,m). PCK@5cm results illustrate the same trend. Pairwise comparisons statistically indicate that MAEPose outperforms baselines across folds.

\textbf{\textit{mmMove:}} The superior performance of MAEPose remains on mmMove, a dataset of similar action categories collected in another larger room performed by non-overlapping participants. MAEPose reaches the overall best MPJPE (0.0307\,m), with an improvement of over 20.3\% over the best baseline (PoseFormer). Statistical results reveal that MAEPose consistently outperforms baselines at the LOPO level.

\textbf{\textit{mmYoga:}} This dataset brings challenges for pose estimation precision, as all models' performances are relatively lower than on the other two datasets. This is due to the fact that this dataset consists of more complex and challenging action categories, requiring participants to move gently with lower motion speed while keeping the body torso controlled. However, we observed during data collection that participants had less control over their body torso. This natural challenge results in the dataset having less repeatable movement instances with more variations, hence increasing the complexity of the dataset. But again, the trend remains constant, where MAEPose achieves an MPJPE of 0.0419\,m (6.3\% improvement over the best baseline, PoseFormer) and PCK@5cm of 75.4\%. Statistical significance demonstrates MAEPose's advantages over baselines when facing this complex dataset.

Across all three datasets, MAEPose consistently demonstrates superior performance over baselines based on LOPO cross validation. This reveals that its unique design of the self-supervised pre-trained mmWave video feature encoder with its multi-frame heatmap decoder improves mmWave-based pose estimation across different indoor environments, participant groups and action complexities.

\subsection{Action-Level Analysis}
\begin{figure*}[t]
  \centering
  \begin{subfigure}[t]{0.32\textwidth}
    \centering
    \includegraphics[width=\textwidth]{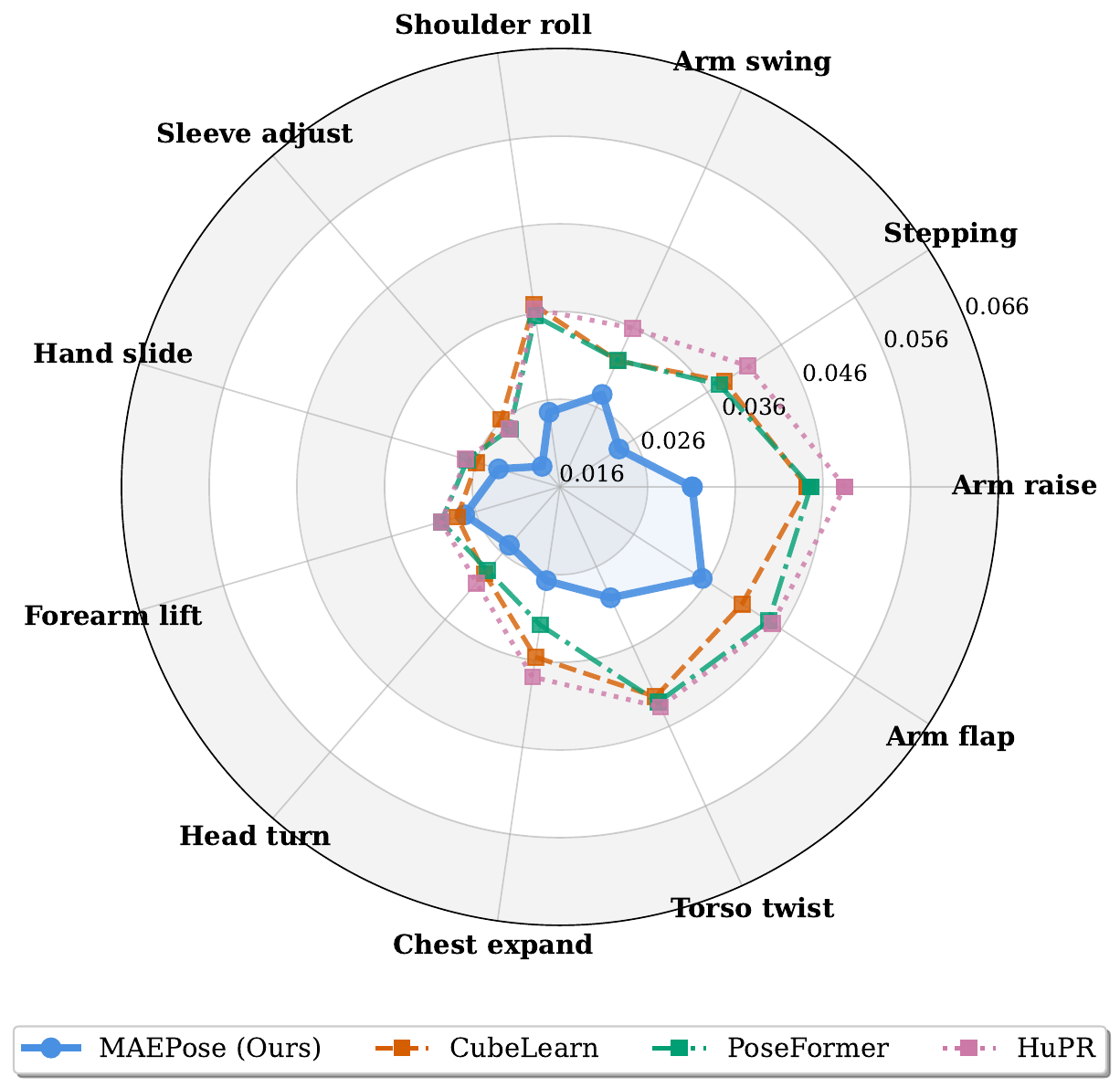}
    \caption{\textit{mmTryOn} (11 actions)}
    \label{fig:radar_mmTryOn}
  \end{subfigure}
  \hfill
  \begin{subfigure}[t]{0.32\textwidth}
    \centering
    \includegraphics[width=\textwidth]{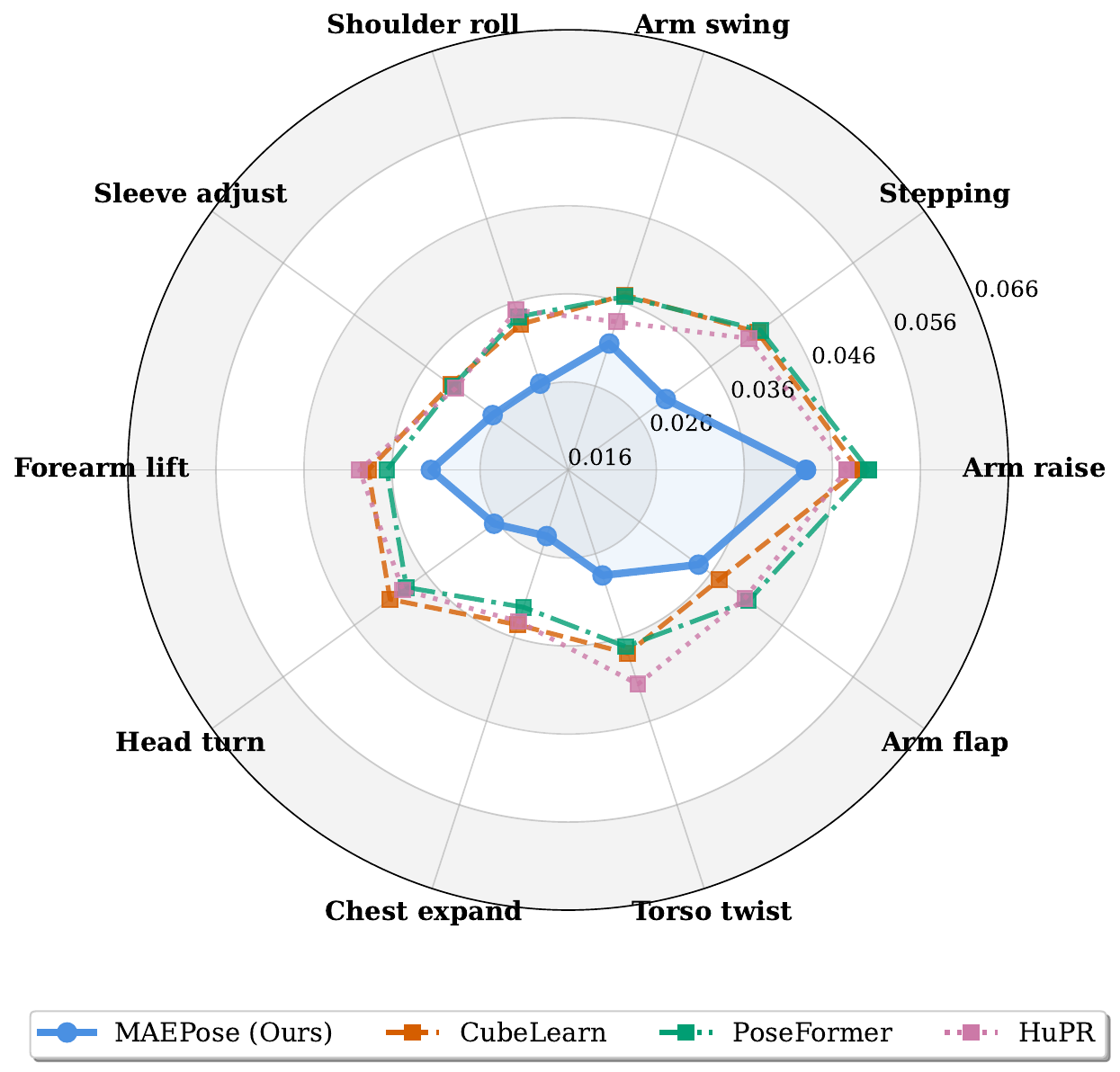}
    \caption{\textit{mmMove} (10 actions)}
    \label{fig:radar_mmMove}
  \end{subfigure}
  \hfill
  \begin{subfigure}[t]{0.32\textwidth}
    \centering
    \includegraphics[width=\textwidth]{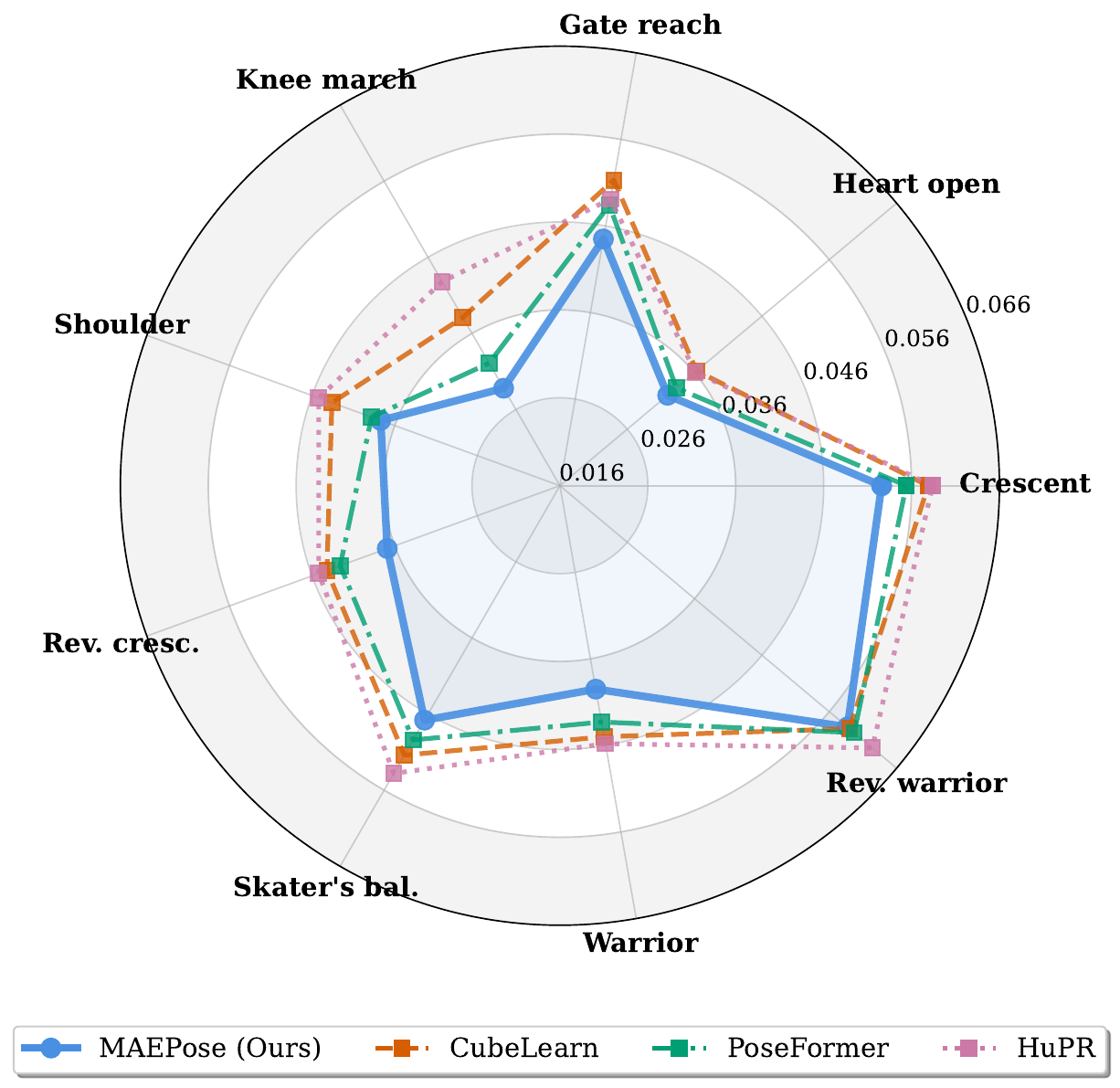}
    \caption{\textit{mmYoga} (9 actions)}
    \label{fig:radar_mmYoga}
  \end{subfigure}
  \caption{Action-level model performances visualization across three datasets. Results are based on MPJPE where lower values indicate better pose estimation accuracy.}
  \label{fig:per_action}
\end{figure*}

We further report model performance across all datasets at the action level. Figure~\ref{fig:per_action} shows the per-action MPJPE for MAEPose and all baselines, plotted on radar charts with a unified radial scale for fair cross-dataset comparison. The plots illustrate how pose estimation difficulty varies across actions, where models tend to predict more accurately on large-movement actions. Specifically, MAEPose (coloured in blue) achieves the lowest error on dynamic actions such as \textit{Stepping} ($-$28\%) and \textit{Chest expand} ($-$26\%) on \textit{mmMove}, and \textit{Rev.\ crescent} ($-$13\%) and \textit{Knee march} ($-$10\%) on \textit{mmYoga}. For slow or subtle-movement actions, although MAEPose maintains its lead, the gap narrows, as such low-motion actions create fewer Doppler variations in the mmWave signal, increasing the discriminative difficulty.

\subsubsection{Qualitative Pose Estimation Analysis}
\begin{figure}[]
  \centering
  \includegraphics[width=0.95\columnwidth]{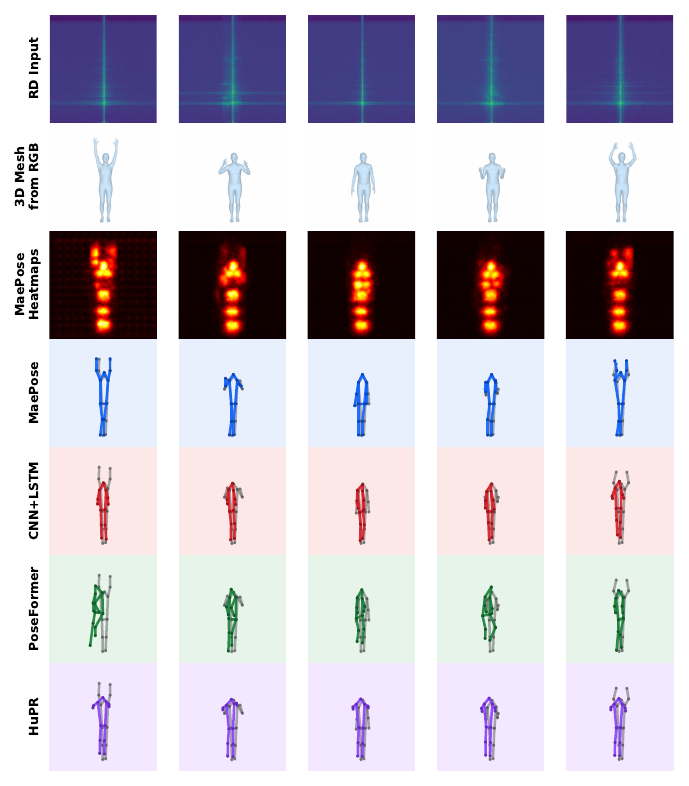}
  \caption{A visualization of pose estimation result from all models on the ``arm raise'' action.}
  \label{fig:skeleton_vis}
  \vspace{-1em}
\end{figure}

Figure~\ref{fig:skeleton_vis} presents a qualitative pose estimation result from all models' predictions. The instance is sampled from the ``arm raise'' action in the \textit{mmTryOn} dataset. From top to bottom: the mmWave Range-Doppler frames (evenly select 5 of 20 frames to simplify visualization), 3D body mesh extracted from the corresponding RGB frames (for pose reference only, replacing the participant's RGB images due to privacy concern), the heatmaps predicted by MAEPose, MAEPose predicted skeletons via argmax calculation over heatmaps, and the three baseline results. Grey-coloured ground truth skeleton is plotted over each prediction row for direct comparison.

\subsection{Modality Study}
\label{sec:modality_study}
As introduced in \S\ref{sec:sensing_principle}, mmWave radar raw data can produce multiple modalities: Range-Doppler (RD) encodes per-range radial velocity, and Range-Angle (RA) encodes per-range azimuth, capturing information from different perspectives. 

To our knowledge, no prior work has systematically compared these modalities as video-format inputs for pose estimation. We, therefore, conduct a study to understand how varying input modality affects model pose estimation accuracy, evaluated consistently based on LOPO cross validation. Specifically, we compare model performances across three variations: single-modality inputs of Range-Doppler or Range-Azimuth, and dual-inputs fusing RD and RA as two parallel input streams, under a unified architecture where only the input changes.

To unify the modality input across models, for single-modality video inputs (RD or RA), all models can directly adapt to the chosen video stream, where baselines take either RD or RA video as input to regress skeleton keypoint coordinates, while MAEPose pre-trains and fine-tunes on the same input modality for heatmap generation and applies argmax for skeleton outputs. For dual-modality inputs, we duplicate the Conv3d input-feature extractor into two parallel branches to process both RD and RA video streams simultaneously. The extracted features from both streams are then concatenated with cross attention before being passed to the latent layers for predicting skeletons.

Table~\ref{tab:main_results} presents model performances under all modality configurations. \textbf{Overall, a consistent performance trend emerges across three datasets: RD\,$\approx$\,Dual\,$\gg$\,RA.} For MAEPose, RD alone matches or slightly outperforms Dual-based ones across all datasets, however RA-based ones show estimation accuracies roughly 20\% lower than the other two in MPJPE. For \textit{mmTryOn}, RD achieves 0.0268\,m, followed by Dual-based ones of 0.0265\,m ($-$1.1\%), whereas RA drops to 0.0345\,m (error$+$28.7\%). For \textit{mmMove}, RD (0.0307\,m) remains the best modality, slightly better than the Dual models (0.0313\,m, error$+$2.0\%), but significantly better than RA (0.0381\,m, error$+$24.1\%). Regarding \textit{mmYoga}, RD (0.0419\,m) is again the best, with Dual at 0.0433\,m ($+$3.3\%) and RA at 0.0527\,m ($+$25.8\%).

The results across all three datasets reveal that the RD video stream alone captures the motion information needed for mmWave-based pose estimation. In contrast, RA alone provides notably less motion-related information compared to RD. Therefore, fusing both RD and RA as dual stream does not bring a significant improvement compared to RD alone, while doubling the cost. We attribute this to the fact that the two mmWave modality streams carry different dimensions of meaning: RD presents the motion-related features through the Doppler bins while RA describes the azimuth of the reflecting targets rather than motion-level granularity, making it less informative for pose estimation tasks.

\subsection{Zero-shot Real-World Evaluation}
A challenge for mmWave sensing is the characteristic variations between the controlled settings (source domain, single-person clear environment) and the real-world deployment (target domain, interference from unexpected reflection targets). For instance, in real-world settings, bystanders, the irrelevant persons who appear and move randomly in the radar's field of view, can introduce unexpected multi-path reflections, interference and overlapping on the Doppler signatures, resulting in domain variations for the model.

To investigate the robustness of MAEPose and baselines when facing such domain variations, we collect an additional dataset where a participant is performing actions and, at the same time, a bystander appears and walks freely with upper-body random movements around the participant (move in between the participant and the mmWave sensor). Actions are performed by the same group of 9 participants recruited for the \textit{mmMove} dataset. Action categories are also reused. Data is collected in the same studio room but with furniture (tables and chairs) relocated, resulting in 6,040 radar frames. We use this additional dataset as the testing set to directly test models' performance through LOPO cross validation without any re-training/fine-tuning adaptation, which makes this a purely zero-shot evaluation.

\begin{table}[tb]
\centering
\caption{Performances under zero-shot real-world situation. $p$-value: Bonferroni-adjusted MPJPE-based pairwise test vs MAEPose.}
\label{tab:robustness}
\footnotesize
\setlength{\tabcolsep}{3pt}
\renewcommand{\arraystretch}{1.1}
\begin{tabular}{@{}lccc@{}}
\toprule
Method & MPJPE\,$\downarrow$ & PCK\,$\uparrow$ & $p$-value \\
\midrule
CubeLearn      & \second{.0404{\tiny$\pm$.009}}          & \second{76.7{\tiny$\pm$11}}          & .0222 \\
PoseFormer     & .0415{\tiny$\pm$.010}                   & 76.3{\tiny$\pm$8}                    & .0444 \\
HuPR           & .0440{\tiny$\pm$.012}                   & 75.9{\tiny$\pm$8}                    & .0740 \\
MAEPose (Ours) & \best{.0328{\tiny$\pm$.011}} \gain{18.8} & \best{84.2{\tiny$\pm$9}} \gain{9.8} & ---   \\
\bottomrule
\end{tabular}
\end{table}

Table~\ref{tab:robustness} reports model performances under the zero-shot real-world setting where interference and noise brought by a bystander walking freely while participant performing postures. All models, to different extents, show performance drops due to the interference noise. However, MAEPose demonstrates the highest robustness, still achieving the best accuracy of 0.0328\,m with only a 6.5\% error increase compared to its controlled-setting performance, and remaining 18.8\% lower in MPJPE than the best baseline (CubeLearn: 0.0404\,m). We attribute this robustness to MAEPose's spatiotemporal representation learned during pre-training on clean radar video, where the encoder learns to embed key motion patterns across spacetime. As a result, even when some frames are partially overwhelmed by unexpected reflections, MAEPose can still rely on spatiotemporal context conveyed from other frames to extract robust motion-related features for pose estimation. In contrast, baselines lack this spatiotemporal prior knowledge and become more sensitive to per-frame noise.

\section{Ablation Study}
\subsection{Impact of Pre-training}
\label{sec:ablation_pre-train}

To investigate the contribution of self-supervised pre-training of MAEPose, we train an identical model but with random initialization without the masked autoencoding pre-training task (directly starting at stage 2, as presented in Figure~\ref{fig:MAEPose_model}). We report MAEPose trained via random-init versus the two-stage pipeline in Table~\ref{tab:ablation_pt} for all datasets.

It is clear that the spatiotemporal representation learnt through MAEPose pre-training brings significant improvements for the downstream pose estimation tasks across all datasets: $-$20.1\% MPJPE on \textit{mmTryOn}, $-$18.8\% on \textit{mmMove}, and $-$41.2\% on \textit{mmYoga}. This indicates that self-supervised pre-training helps models to gain the generalized mmWave video representation in advance, while during the fine-tuning stage, the model focuses on aligning the gap between the generalized knowledge and diverse motion patterns with labels in a supervised manner, leading to higher pose estimation accuracies.

\begin{table}[tb]
\centering
\caption{Impact of self-supervised pre-training on MAEPose performances. $p$-value: MPJPE-based pairwise test vs Pre-trained.}
\label{tab:ablation_pt}
\footnotesize
\setlength{\tabcolsep}{4pt}
\renewcommand{\arraystretch}{1.1}
\begin{tabular}{@{}llccc@{}}
\toprule
Dataset & Config & MPJPE (m)\,$\downarrow$ & PCK (\%)\,$\uparrow$ & $p$-value \\
\midrule
\multirow{2}{*}{\textit{mmTryOn}} & Random Init & .0335{\tiny$\pm$.005} & 87.0{\tiny$\pm$5} & .0001 \\
 & Pre-trained & \best{.0268{\tiny$\pm$.004}} \gain{20.1} & \best{91.1{\tiny$\pm$5}} \gain{4.7} & --- \\
\midrule
\multirow{2}{*}{\textit{mmMove}} & Random Init & .0379{\tiny$\pm$.010} & 76.8{\tiny$\pm$17} & .0004 \\
 & Pre-trained & \best{.0307{\tiny$\pm$.008}} \gain{18.8} & \best{85.3{\tiny$\pm$11}} \gain{11.1} & --- \\
\midrule
\multirow{2}{*}{\textit{mmYoga}} & Random Init & .0712{\tiny$\pm$.004} & 43.7{\tiny$\pm$5} & \(<\)\!.0001 \\
 & Pre-trained & \best{.0419{\tiny$\pm$.005}} \gain{41.2} & \best{75.4{\tiny$\pm$5}} \gain{72.5} & --- \\
\bottomrule
\end{tabular}
\end{table}

\subsubsection{Qualitative Video Mask Reconstruction Analysis}
\begin{figure}[tb]
  \centering
  \includegraphics[width=0.9\linewidth]{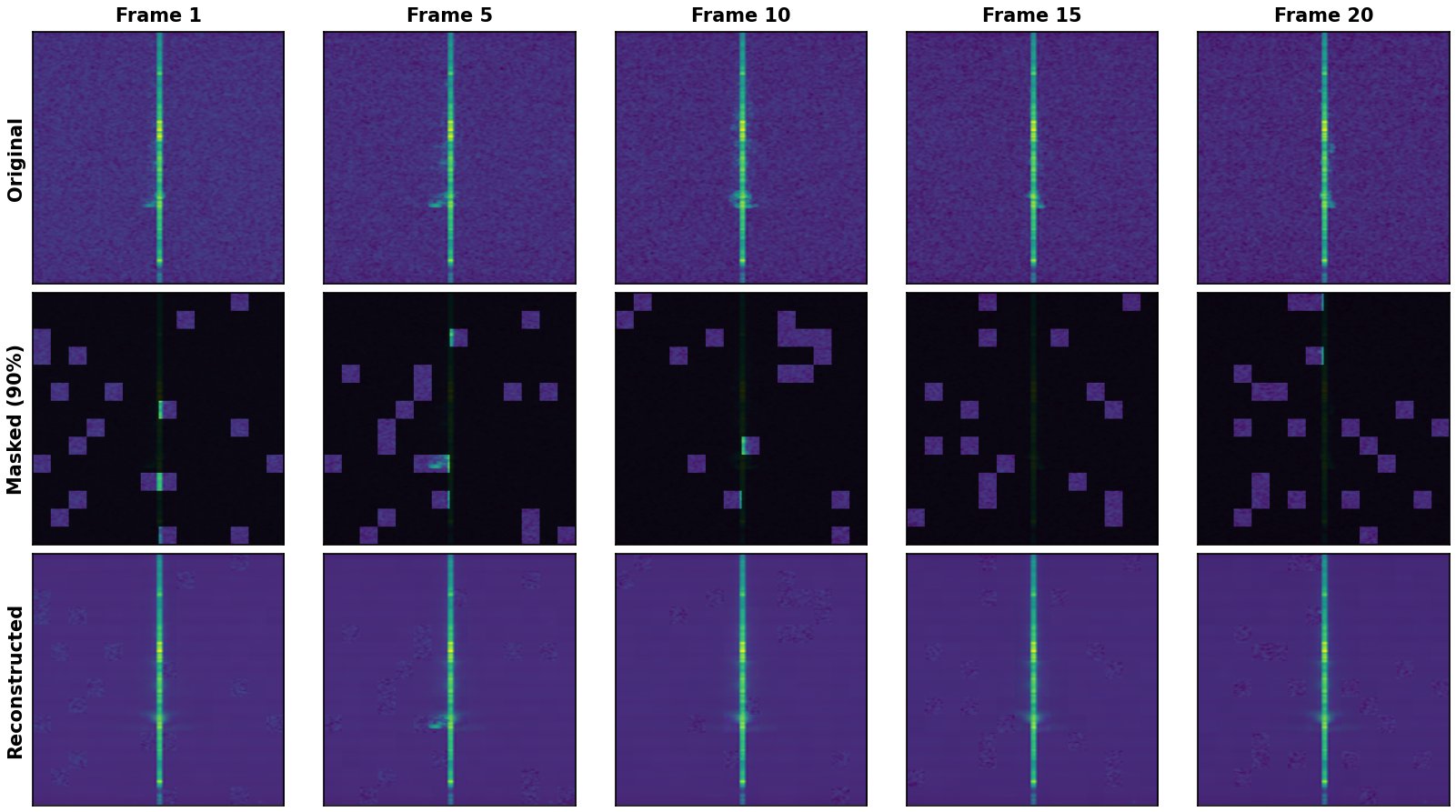}
  \caption{Qualitative MAEPose reconstruction results on an unseen RD video clip. Top: original RD frames. Middle: 90\% masked input (only 10\% patches visible for model inputs). Black patches represent those masked ones, while coloured patches represent what model will receive. Bottom: reconstructed frames from MAEPose reconstruction decoder.}
  \label{fig:mae_reconstruction}
  \vspace{-1em}
\end{figure}

To qualitatively investigate the spatiotemporal understanding capability of MAEPose's video encoder, we visualize MAEPose's masked video reconstruction results in Figure~\ref{fig:mae_reconstruction}. The hypothesis is that if giving an unseen mmWave video sample with 90\% of the patches masked at the input level, the pre-trained encoder can reconstruct the correct micro-Doppler pattern across frames, it indicates that the spatiotemporal representation has been learnt during the masked self-supervised learning process. From top to bottom, Figure~\ref{fig:mae_reconstruction} shows: a sequence of mmWave raw RD frames, the randomly masked frames (what the model actually sees), and the reconstructed frames from MAEPose's video encoder.

Based on the reconstruction results, we observe that the Doppler patterns at the frame level are correctly reconstructed, even when the input masked patches retain very few of the original Doppler features. It recovers the spatiotemporal structure of those Doppler-highlighted pixels while, interestingly, automatically reducing its focus on reconstructing the noisy background which is less informative in the sparse mmWave videos. When the MAEPose encoder deals with a sequence of RD frames across spacetime, the reconstruction process leverages the learnt high-level generalized knowledge to embed and reconstruct the masked patches with higher learnt attention over those motion-related pixels. The reconstruction quality demonstrates the effectiveness of MAEPose's pre-training, showing that such representation is learnt during pre-training, therefore contributing to its downstream pose estimation.

\subsection{Impact of Decoder Architecture}
To investigate the impacts of MAEPose's decoder architecture on pose estimation performance, we include two classic projection head architectures used for pose estimation, replacing the heatmap decoder with: i) an MLP (simple fully-connected layer), and ii) a GCN (Graph Convolutional Network). Both of these two decoders process the extracted features from the same pre-trained encoder to directly regress skeleton joints. Evaluation is consistently based on LOPO cross validation across three datasets.

\begin{table}[tb]
\centering
\caption{MAEPose performances with different decoder architectures. $p$-value: Bonferroni-adjusted MPJPE-based pairwise test vs Heatmap.}
\label{tab:ablation_decoder}
\footnotesize
\setlength{\tabcolsep}{4pt}
\renewcommand{\arraystretch}{1.1}
\begin{tabular}{@{}llccc@{}}
\toprule
Dataset & Decoder & MPJPE (m)\,$\downarrow$ & PCK (\%)\,$\uparrow$ & $p$-value \\
\midrule
\multirow{3}{*}{\textit{mmTryOn}} & MLP & .0403{\tiny$\pm$.010} & 80.1{\tiny$\pm$6} & .0234 \\
 & GCN & .0490{\tiny$\pm$.013} & 74.2{\tiny$\pm$7} & .0234 \\
 & Heatmap & \best{.0268{\tiny$\pm$.004}} \gain{33.5} & \best{91.1{\tiny$\pm$5}} \gain{13.7} & --- \\
\midrule
\multirow{3}{*}{\textit{mmMove}} & MLP & .0378{\tiny$\pm$.006} & 79.5{\tiny$\pm$8} & .0555 \\
 & GCN & .0384{\tiny$\pm$.007} & 76.8{\tiny$\pm$10} & .0279 \\
 & Heatmap & \best{.0308{\tiny$\pm$.008}} \gain{18.6} & \best{85.3{\tiny$\pm$11}} \gain{7.3} & --- \\
\midrule
\multirow{3}{*}{\textit{mmYoga}} & MLP & .0537{\tiny$\pm$.011} & 55.9{\tiny$\pm$15} & .0178 \\
 & GCN & .0608{\tiny$\pm$.011} & 45.4{\tiny$\pm$15} & .0010 \\
 & Heatmap & \best{.0419{\tiny$\pm$.005}} \gain{22.0} & \best{75.4{\tiny$\pm$5}} \gain{34.9} & --- \\
\bottomrule
\end{tabular}
\end{table}

Table~\ref{tab:ablation_decoder} reports the results. The heatmap decoder outperforms both MLP and GCN decoders on all three datasets, where the MLP decoder's error increases by 19--34\% and GCN by 20--45\% in MPJPE. This is likely because both MLP and GCN collapse the rich spatiotemporal information into a compact representation and force the projection head to regress skeleton coordinates directly, limiting it to regress skeleton positions without spatiotemporal correspondence.

In contrast, using the heatmap as decoder better preserves the spatiotemporal features learned by the MAEPose video encoder, rather than discarding such rich information by narrowing the learning process to the downstream task as in the baselines. Furthermore, the heatmap decoder maintains explicit correspondence between feature patterns and skeleton heatmap locations, through its unified visual formulation of both input and output. This allows the model to regard each joint as a spatial confidence map, which is better suited to pose localization, rather than collapsing it through pooling before the downstream task, leading to more accurate joint localization.

\section{Conclusion}

We presented MAEPose, a self-supervised spatiotemporal learning pose estimation framework that operates directly on mmWave spectrogram video streams. MAEPose learns the generalized spatiotemporal representation via masked autoencoding over the unlabelled radar data, followed by a multi-frame heatmap decoder that preserves spatial correspondence for skeleton keypoints prediction. We evaluate MAEPose with baseline comparisons across three datasets spanning diverse participant groups, action categories and indoor environments. MAEPose shows superior performance over the SOTA baselines under leave-one-person-out cross-validation, and demonstrates its robust performance under zero-shot real-world interference. Ablation studies reveal that both the self-supervised pre-training and the heatmap decoder contribute to MAEPose's performance. These results demonstrate the promise of MAEPose on learning spatiotemporal representations for human pose estimation from mmWave spectrogram video. Future work aims to improve cross-environment generalization by adapting MAEPose to new environments through fine-tuning with a small amount of labelled target-room data to reduce deployment costs.

\bibliographystyle{ACM-Reference-Format}
\bibliography{references}

\end{document}